\pdfoutput=1

\documentclass[11pt]{article}

\usepackage[preprint]{acl}

\usepackage{times}
\usepackage{latexsym}

\usepackage[T1]{fontenc}
\usepackage[utf8]{inputenc}
\usepackage{microtype}
\usepackage{inconsolata}
\usepackage{graphicx}
\usepackage{tcolorbox}

\definecolor{myblue}{RGB}{0, 0, 255} 
\definecolor{mydarkblue}{RGB}{0, 0, 139} 

\usepackage{url}            
\usepackage{siunitx}
\usepackage{tikz}
\usepackage{wasysym}
\usepackage{booktabs}       
\usepackage{amsfonts}       
\usepackage{nicefrac}       
\usepackage{microtype}      
\usepackage{adjustbox}

\usepackage{graphicx}
\usepackage{tabularx}
\usepackage{subcaption}
\usepackage{multicol}

\usepackage{amsthm}
\usepackage{algorithm}
\usepackage{algorithmic}
\usepackage{dsfont}

\usepackage{multirow}
\usepackage{bm}
\usepackage{colortbl}
\usepackage{wrapfig}
\usepackage{enumitem}
\usepackage{soul}
\usepackage{makecell}
\usepackage{thm-restate}

\usepackage[capitalize,noabbrev,nameinlink]{cleveref}

\makeatletter 
\newcommand{\fixed@sra}{$\vrule height 2\fontdimen22\textfont2 width 0pt\rightarrow$}
\newcommand{\shortarrow}[1]{%
  \mathrel{\text{\rotatebox[origin=c]{\numexpr#1*45}{\fixed@sra}}}
}
\makeatother

\definecolor{myTeal}{RGB}{55,147,149}
\definecolor{myBluesss}{RGB}{31,117,211}

\definecolor{darkred}{rgb}{0.80, 0.0, 0.0}
\definecolor{royalblue}{rgb}{0.0549, 0.9118, 0.9224} 
\definecolor{forestgreen}{rgb}{0.1333, 0.5451, 0.1333} 
\definecolor{grn}{rgb}{0.1, 0.6, 0.1}
\definecolor{mgt}{rgb}{0.7, 0.3, 0.7}
\definecolor{chamoisee}{rgb}{0.63, 0.47, 0.35}
\definecolor{purp}{rgb}{0.65, 0.16, 0.65}
\definecolor{alizarin}{rgb}{0.82, 0.1, 0.26}
\definecolor{alizarin2}{rgb}{0.9, 0.1, 0.26}
\definecolor{azure(colorwheel)}{rgb}{0.0, 0.5, 1.0}
\definecolor{brown}{rgb}{0.65, 0.16, 0.16}
\definecolor{lblue}{rgb}{0, 0.2, 0.8}
\definecolor{orange}{rgb}{1.0, 0.5, 0.0}
\newcommand{\ggreen}[1]{{\color{myTeal}{#1}}}
\newcommand{\bblue}[1]{{\color{myBluesss}{#1}}}
\newcommand{\rred}[1]{{\color{alizarin2}{#1}}}

\definecolor{darkblue-purple}{RGB}{48, 35, 122}
\definecolor{cyan}{RGB}{0, 100, 138}
\definecolor{darkgreen}{rgb}{0.0, 0.5, 0.0}
\definecolor{darkyellow}{rgb}{0.8, 0.6, 0.1}
\definecolor{myblue}{RGB}{0, 0, 255} 
\definecolor{mydarkblue}{RGB}{0, 0, 139} 

\theoremstyle{plain}

\newtheorem{observation}{Observation}
\theoremstyle{definition}

\theoremstyle{remark}

\title{Self-Training Large Language Models with Confident Reasoning}

\author{%
  Hyosoon Jang$^1$, Yunhui Jang$^2$, Sungjae Lee$^1$, Jungseul Ok$^1$, Sungsoo Ahn$^2$ \\
  $^1$POSTECH\quad$^2$KAIST \\
  \texttt{\{hsjang1205,sungjaelee25,jungseul\}@postech.ac.kr,} \\ \texttt{\{yunhuijang,sungsoo.ahn\}@kaist.ac.kr} 
}

\begin{document}

\maketitle

\begin{abstract}
Large language models (LLMs) have shown impressive performance by generating reasoning paths before final answers, but learning such a reasoning path requires costly human supervision. To address this issue, recent studies have explored self-training methods that improve reasoning capabilities using pseudo-labels generated by the LLMs themselves. Among these, confidence-based self-training fine-tunes LLMs to prefer reasoning paths with high-confidence answers, where confidence is estimated via majority voting. However, such methods exclusively focus on the quality of the final answer and may ignore the quality of the reasoning paths, as even an incorrect reasoning path leads to a correct answer by chance. Instead, we advocate the use of reasoning-level confidence to identify high-quality reasoning paths for self-training, supported by our empirical observations. We then propose a new self-training method, \textbf{CORE-PO}, that fine-tunes LLMs to prefer high-\textbf{CO}nfidence \textbf{RE}asoning paths through \textbf{P}olicy \textbf{O}ptimization. Our experiments show that CORE-PO improves the accuracy of outputs on four in-distribution and two out-of-distribution benchmarks, compared to existing self-training methods.
\end{abstract}
\section{Introduction} 

Large language models (LLMs) have shown impressive performance across various tasks by generating reasoning paths before yielding the final answer \citep{wei2022chain,kojima2022large,zhang2023automatic}. However, the potential for improving reasoning through supervision is limited by the scarcity of high-quality data with human-annotated or ground-truth labels. To address this issue, recent studies have proposed \textit{self-training methods} for LLMs, which leverage pseudo-labels generated by the LLMs themselves and require only the input questions~\citep{huang2023large,kumar2024self-rewarding,prasad2024self,zhang2024chain,zhang2024self,ranaldi2024self,zuo2025ttrltesttimereinforcementlearning}.\footnote{We refer to self-training as a scheme that requires only the input questions, without labels or external models.} At a high level, these methods fine-tune the base LLMs to prefer high-quality outputs identified through self-assessment strategies in inference-time scaling techniques, e.g., self-consistency~\citep{wang2023selfconsistency}, tree-of-thoughts~\citep{yao2023tree}, and self-refinement~\citep{madaan2023selfrefine}.

As a representative approach, confidence-based methods have improved the reasoning capabilities by training LLMs to prefer reasoning paths associated with high-confidence answers \citep{huang2023large, prasad2024self, zhang2024self,zuo2025ttrltesttimereinforcementlearning}. These methods are motivated by the observation that the answers with high confidence scores tend to yield high accuracy \citep{wang2023selfconsistency,taubenfeld2025confidence}, and thus assume that reasoning paths leading to such answers are reliable. Specifically, \citet{huang2023large, prasad2024self} and \citet{zhang2024self} estimate the confidence in answers using self-consistency scores measured via majority voting, and then fine-tune LLMs to prefer reasoning paths associated with high-confidence answers.

\begin{figure*}[t]
\centering
\;\;\includegraphics[width=0.9\linewidth]{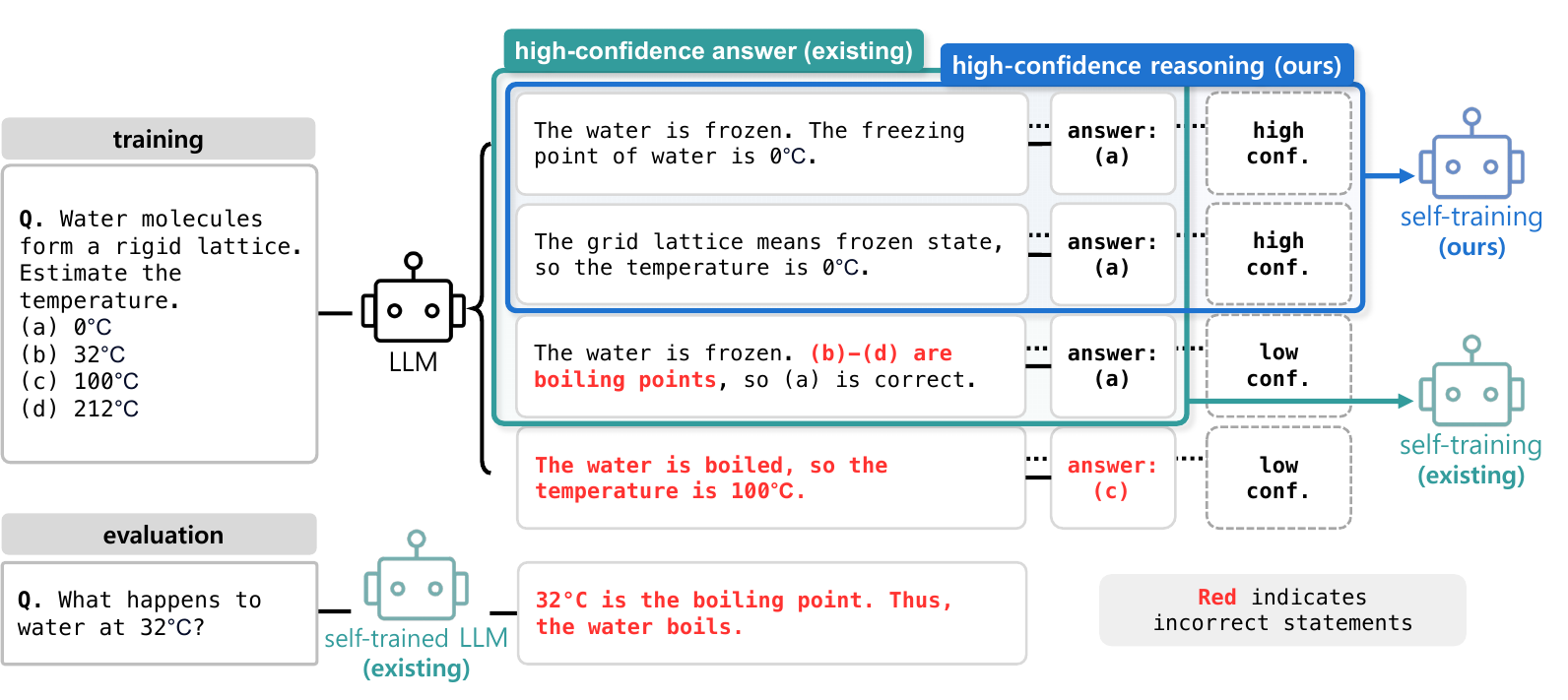}
\vspace{-0.1in}
\caption{\textbf{Limitations in existing confidence-based self-training methods.} Existing self-training methods prefer \ggreen{reasoning paths associated with a high-confidence answer (a)}, estimated via majority voting. However, they fail to capture the \rred{errors in their third reasoning path}, even though the answer is correct. As a result, they can degrade the reasoning capabilities of the LLM, e.g., preferring \rred{``(b)-(d) are boiling points''} can lead to \rred{``$32\,^{\circ}\mathrm{C}$ is the boiling point''}, as shown in below. In contrast, our method measures reasoning-level confidence (as depicted by the dashed line) and fine-tunes LLMs to prefer \bblue{high-confidence reasoning paths} that yield correct statements.}\label{fig:intro}
\vspace{-0.1in}
\end{figure*}


In this work, we argue that existing confidence-based self-training methods exclusively focus on \textit{answer-level confidence} and may ignore the quality of the reasoning path. In practice, as illustrated in \Cref{fig:intro} and observed in \Cref{observation}, LLMs often generate incorrect reasoning paths that lead to high-confidence answers, even when those answers are correct \citep{lanham2023measuring,zhang2024restmcts}. Consequently, LLMs may learn to prefer incorrect reasoning paths associated with high-confidence answers, which degrade their reasoning capabilities. This pitfall highlights the necessity of incorporating reasoning-aware confidence measures into the self-training of LLMs to better identify high-quality reasoning paths.


To this end, we propose incorporating \textit{reasoning-level confidence} into a confidence-based self-training method. As illustrated in \Cref{fig:intro}, our method evaluates the correctness of reasoning by estimating confidence in the reasoning paths rather than relying solely on answer-level confidence. This is motivated by \Cref{observation2}, which shows that outputs with higher reasoning-level confidence exhibit fewer errors, aligned with prior findings that such confidence is useful for identifying high-quality outputs \citep{becker2024cyclesthoughtmeasuringllm,wan-etal-2025-reasoning,taubenfeld2025confidence}. 





We then propose \textbf{CORE-PO}, a method that fine-tunes LLMs to prefer high-\textbf{CO}nfidence \textbf{RE}asoning paths using \textbf{P}olicy \textbf{O}ptimization. {To be specific, we estimate the reasoning-level confidence using $\text{P(True)}$ \citep{kadavath2022languagemodelsmostlyknow}, measuring the probability that LLM returns ``true'' to the prompt asking whether the reasoning is correct. We consider the two ways to measure $\text{P(True)}$ of reasoning: a monolithic way that assesses the entire reasoning path, and a statement-wise way that computes the average confidence across each step in the reasoning path. Then, we combine reasoning-level confidence with answer-level confidence, and fine-tune LLMs using direct preference optimization~\citep{guo2024direct} to prefer high-confidence outputs.} 

In our experiments, we apply our self-training method to four arithmetic or scientific reasoning benchmarks: GSM8K \citep{cobbe2021training}, ARC-Challenge \citep{clark2018think}, GPQA \citep{rein2023gpqa}, and MATH \citep{hendrycksmath2021}. We also consider two external benchmarks, CRUXEval~\citep{pmlr-v235-gu24c} and Game-of-24 \citep{lile2025game24}, to evaluate the generalization capabilities on out-of-distribution tasks. Our method improves the accuracy of outputs on both in-distribution and out-of-distribution tasks by enhancing reasoning quality, compared to existing self-training approaches.


To conclude, our contributions can be summarized as follows:
\begin{itemize}[topsep=-1.0pt,itemsep=1.0pt,leftmargin=3.5mm]
\item {We identify a limitation of existing confidence-based self-training methods: they rely solely on \textit{answer-level confidence}, which may fail to capture the errors in reasoning.}
\item {We propose a new self-training method that incorporates \textit{reasoning-level confidence} to better identify reasoning paths with fewer errors.}
\item {Through extensive evaluation, we show that our method improves answer accuracy and reduces errors in reasoning compared to existing approaches on both in-distribution and out-of-distribution reasoning benchmarks.}
\end{itemize}

\begin{figure*}[t]
\setcounter{figure}{2}  
\centering
\;\;\includegraphics[width=0.9\linewidth]{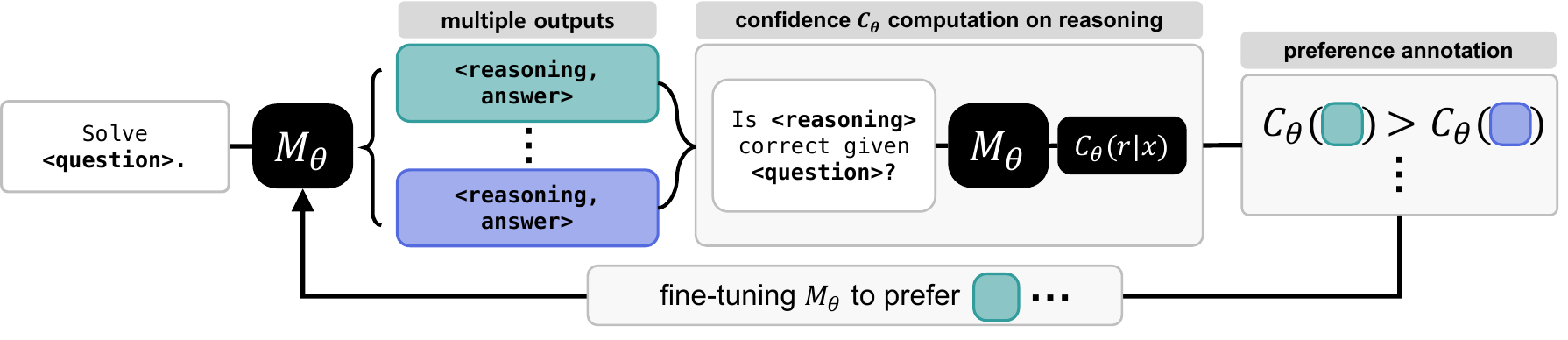}
\vspace{-.07in}
\caption{\textbf{Overview of CORE-PO.} The LLM $M_{\theta}$ generates multiple outputs, each consisting of a reasoning and an answer $s = [r, a]$ for a given question. Next, we measure the reasoning-level confidence $C_{\theta}(r|x)=\text{P(True)}$ for each reasoning path. Then, we fine-tune the LLM to prefer high-confidence reasoning paths.} \label{fig:method}
\vspace{-.13in}
\end{figure*}

\section{Background: Self-Training LLMs} 

In this section, we describe the preliminaries of existing self-training methods for large language models~(LLMs). We describe additional related works in \Cref{sec:related}.

\noindent \textbf{Notation.} Let $x$ denote a question provided to an LLM $M_{\theta}$. For a given question $x$, we assume that the LLM outputs a sequence $s = [r, a]$, where $r$ represents the reasoning path and $a$ is the final answer induced from this reasoning path. In this paper, we are particularly interested in the model’s confidence in its generated sequences. We denote the model’s confidence score (or uncertainty estimate) on a given statement by $C_{\theta}(\cdot)$, which can be computed through various existing approaches such as self-consistency \citep{wang2023selfconsistency}, semantic entropy \citep{kuhnsemantic}, or other uncertainty quantification techniques. 


\subsection{Existing works on self-training} 

Recent studies have shown that LLMs can self-improve through fine-tuning with pseudo-labels, e.g., the preference, generated by themselves. The key idea of these approaches is to transfer the performance gains from inference-time scaling methods such as self-consistency \citep{wang2023selfconsistency}, tree-of-thoughts \citep{yao2023tree}, or self-refinement \citep{madaan2023self}, into training-time improvements through fine-tuning. These inference-time methods typically generate multiple candidate outputs and select high-quality outputs based on self-assessments, e.g., confidence estimation \citep{wang2023selfconsistency}. Extending this approach, self-training methods fine-tune the LLMs to prefer outputs assessed as high-quality by the models themselves, leading to improved performance across various tasks \citep{huang2023large,yuan2024icml,prasad2024self,zhang2024self,zuo2025ttrltesttimereinforcementlearning}. Among these, several methods \citep{huang2023large,prasad2024self,zuo2025ttrltesttimereinforcementlearning} leverage confidence scores, providing evidence that confidence-guided supervision can serve as a powerful training signal.


\subsection{Existing works on confidence-based self-training}

Recent self-training methods for LLMs aim to improve the reasoning capabilities by rewarding a reasoning path $r$ that leads to an answer $a$ with a high confidence score $C_{\theta}(a|x)$ \citep{huang2023large, prasad2024self, zhang2024self}. These methods build on the observation that such an answer yields higher accuracy \citep{wang2023selfconsistency}, and thus assume that the reasoning path $r$ leading to such an answer is reliable. Specifically, they estimate the confidence score for an answer $a$ using majority voting over multiple generated answers $a^1, \ldots, a^N$, i.e., $C_{\theta}(a|x) = \frac{1}{N}\sum_{i=1}^N \mathds{1}[a = a^i]$, following the concept of self-consistency \citep{wang2023selfconsistency}. Next, they fine-tune the LLMs using reinforcement learning to prefer reasoning paths with high answer-level confidence $C_{\theta}(a \mid x)$.

%

\section{Method} 


We introduce our confidence-based self-training method for large language models (LLMs) to improve their reasoning capabilities. First, we show (1) how existing confidence-based self-training methods can prefer incorrect reasoning paths, and (2) how incorporating reasoning-level confidence mitigates this issue (\Cref{subsec:motiv}). Next, we describe a method that fine-tunes LLMs to prefer high-\textbf{CO}nfidence \textbf{RE}asoning paths using \textbf{P}olicy \textbf{O}ptimization, coined \textbf{CORE-PO} (\Cref{subsec:our_method}).

\subsection{Motivation for reasoning-level confidence in self-training} \label{subsec:motiv} 

Our motivation stems from the limitations of confidence measures used for existing self-training methods \citep{huang2023large, prasad2024self, zhang2024self}, which evaluate a reasoning–answer pair $[r, a]$ based on the confidence score on the answer $C_{\theta}(a|x)$. Here, we argue that such answer-level confidence may fail to capture the overall quality of the reasoning path (\Cref{fig:intro,observation}), as even an incorrect reasoning path may lead to a correct answer~\citep{lanham2023measuring,zhang2024restmcts}. To remedy this, we advocate the use of reasoning-level confidence $C_{\theta}(r|x)$ as a way to evaluate the quality of reasoning paths~(\Cref{observation2}). 


\begin{figure}
\setcounter{figure}{1}  
\centering
\; \centering \includegraphics[width=0.91\linewidth]{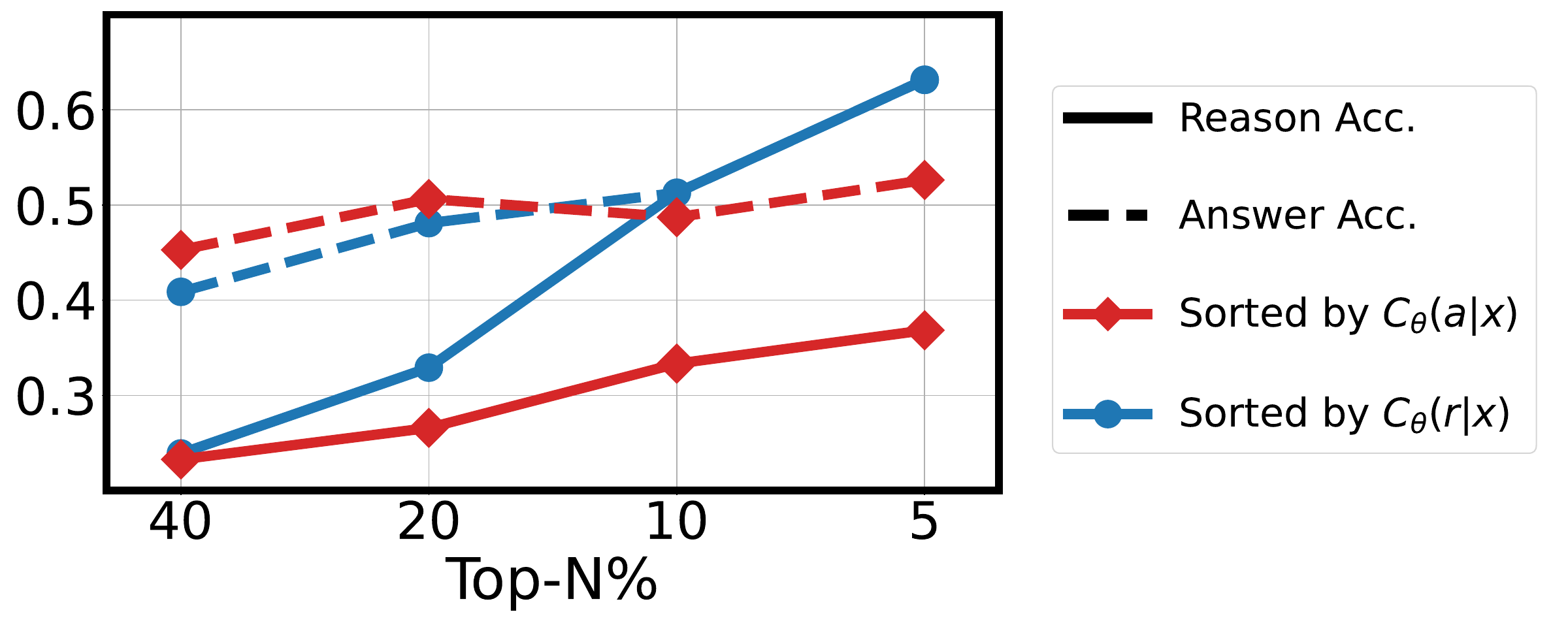}
\vspace{-.1in}
\caption{\textbf{Answer vs. reasoning accuracies.} We use Llama3.1-8B-Instruct \citep{meta2024llama3}. Reasoning-level accuracy coincides more closely with reasoning-level confidence than with answer-level confidence.}\label{fig:motiv}
\vspace{-.1in}
\end{figure}

To support our claim, we conduct an observational experiment using multiple-choice questions in the GPQA dataset \citep{rein2023gpqa}. We generate reasoning paths and assign two types of confidence scores to each path: one based on confidence in the final answer $C_{\theta}(a|x)$ (estimated via majority voting), and the other incorporating confidence in the reasoning path $C_{\theta}(r|x)$.\footnote{We actually consider $C_{\theta}(r,a|x)=C_{\theta}(r|x)C_{\theta}(a|x,r)$ for the later, using the measures described in \Cref{subsec:our_method}.} We describe detailed experimental settings in \Cref{appx:observe}. Then, we evaluate the correctness of the final answers and reasoning paths with accuracy.\footnote{We use \texttt{o4-mini-2025-04-16} \citep{openai2025o4mini} to evaluate correctness of generated reasoning paths.}

Our experiment makes the following observations, which support the use of reasoning-level confidence to evaluate the reasoning path. 
\begin{observation}\label{observation}
\textbf{\upshape Reasoning paths with high answer-level confidence are often incorrect, even when the final answers are correct.} See the gap between the answer-level accuracy (dashed red) and reasoning-level accuracy (solid red) for outputs with high answer-level confidence in \Cref{fig:motiv}. 
\end{observation}
\begin{observation}\label{observation2}
\textbf{\upshape High reasoning-level confidence $C_{\theta}(r|x)$ tends to yield accurate reasoning with fewer errors, coinciding with high answer-level accuracy.} See the accuracy of outputs with high reasoning-level confidence (blue) in \Cref{fig:motiv}.
\end{observation}
The findings in \Cref{observation} hint at the pitfall of existing confidence-based self-training methods, that can train LLMs to prefer incorrect reasoning paths associated with high answer-level confidence by chance. Note that this pitfall also exists in conventional (not self-training) fine-tuning methods that use ground-truth answers and define the reward based on answer-level accuracy \citep{zelikman2022star,trung-etal-2024-reft,deepseekai2025deepseekr1incentivizingreasoningcapability}, as they may assign a positive reward to the incorrect reasoning path that leads to correct answer.\footnote{See the ablation study in \Cref{subsec:abb} for details.}


In response, we propose incorporating reasoning-level confidence $C_{\theta}(r \mid x)$ to identify high-quality reasoning paths for self-training. While confidence scores have primarily been used for the factuality of a single statement, we highlight their utility in evaluating reasoning correctness, supported by \Cref{observation2}. Note that this observation also aligns with the finding in recent inference-time scaling methods~\citep{becker2024cyclesthoughtmeasuringllm,taubenfeld2025confidence}, which improve performance by selecting high-confidence reasoning paths.

\begin{algorithm}[t]
\caption{Self-training LLMs with reasoning-level confidence scores}
\label{alg:algorithm}
\begin{algorithmic}[1]
   \STATE \textbf{Input:} An LLM $M_{\theta}$, a set of questions $\mathcal{X}$
   \STATE Initialize the reference model $M_{\text{ref}}$ using $M_{\theta}$
   \REPEAT 
   \STATE Sample a question $x \sim \mathcal{X}$
   \STATE Sample $\{(r^i, a^i)\}_{i=1}^N \sim M_{\theta}(\cdot \mid x)$
   \STATE Compute $\{C_{{\theta}}(r^i, a^i \mid x)\}_{i=1}^{N}$ 
   \STATE Update $\theta$ to minimize $\mathcal{L}$ in \Cref{eq:objective}
   \UNTIL{convergence}
\end{algorithmic}
\end{algorithm}

\newcommand{\cll}{\cellcolor{blue!4}}

\begin{table*}[t]
\centering
{
\scalebox{0.95}{\begin{tabular}{cccccc}
\toprule
\textbf{Fine-tuning} & \textbf{Decoding} & GSM8K  & ARC-Challenge & GPQA$^{\text{ext}}$ & MATH$^{\text{lv5}}$ \\
\midrule
\midrule
\multirow{4}{*}{Not Applied} & Greedy &  84.2 & 84.5 & 32.4 & 22.6 \\
 & \cll Linguistic & \cll 85.7 & \cll 86.0 & \cll 31.8 & \cll 22.1 \\
& \cll SC & \cll 89.6 & \cll 86.6 & \cll 34.3 & \cll 25.6 \\
& \cll  $\text{P(True}|r,a\text{)}$ & \cll 89.7 & \cll 87.0 & \cll 34.5 & \cll 25.2 \\
\hline
\rowcolor{gray!20}\multicolumn{6}{l}{\textit{Learning from linguistic self-assessment of answer quality}} \\
\multirow{2}{*}{SR-PO \citep{kumar2024self-rewarding}} & Greedy & 85.2 & 86.2 & 34.3 & 19.8 \\ 
& \cll Linguistic & \cll 86.7 & \cll 87.4 & \cll 35.5 & \cll 21.2\\
\hline
\rowcolor{gray!20}\multicolumn{6}{l}{\textit{Learning to prefer reasoning paths with high answer-level confidence $C_{{\theta}}(a|x)$}} \\
\multirow{2}{*}{SC-PO \citep{prasad2024self}} & Greedy & 85.7 & 86.0 & 33.7 & \textbf{25.1} \\
&  \cll SC & \cll 89.7 & \cll 87.5 & \cll 34.5 & \cll 29.4\\
\hline
\rowcolor{gray!20}\rowcolor{gray!20}\multicolumn{6}{l}{\textit{Learning to prefer reasoning with high confidence scores $C_{{\theta}}(r|x)$ \textbf{(ours)}}} \\
\multirow{2}{*}{\textbf{CORE-PO} \textbf{(ours)}} & Greedy & \textbf{86.8} & \textbf{87.5} & \textbf{35.5} & {24.6}  \\
& \cll \cll$\text{P(True}|r,a\text{)}$ & \cll \textbf{90.5} & \cll \textbf{89.2} & \cll \textbf{36.1} & \cll \textbf{29.8} \\
\bottomrule
\end{tabular}
}}
\caption{\textbf{Performance of self-training methods on Llama3.1-8B-Instruct}. \textbf{Bold} indicates the best performance under greedy decoding or \colorbox{blue!5}{inference-time scaling methods} (with sampling of eight outputs). Our method outperforms the considered baselines when applying both greedy decoding and inference-time scaling method.} 
\label{tab:llama}
\end{table*}

\subsection{CORE-PO: Self-training LLMs with reasoning-level confidence} \label{subsec:our_method} 

We describe our self-training method, which learns to prefer high-confidence reasoning paths using policy optimization (CORE-PO). This method involves measuring reasoning-level confidence and training LLMs to prefer high-confidence reasoning paths. We provide an overview of our method in \Cref{fig:method} and \Cref{alg:algorithm}.

To measure the reasoning-level confidence $C_{\theta}(r|x)$, we use $\text{P(True)}$ \citep{kadavath2022languagemodelsmostlyknow} which measures the probability that LLM returns ``true'' to the prompt asking whether the given reasoning $r$ is correct for the given question $x$.\footnote{Note that $\textbf{P(True)}$ has already shown promising results to evaluate the reasoning paths in existing inference-time scaling methods \citep{becker2024cyclesthoughtmeasuringllm,taubenfeld2025confidence}} 
In detail, we consider two ways to measure confidence in a multi-statement reasoning path. First, for the \text{monolithic $\text{P(True)}$}, we measure confidence in one-shot through asking the LLM to check the truthfulness of all the statements at once. Next, for the \text{statement-wise $\text{P(True)}$}, we query the LLM for each statement to check the truthfulness, then average over the confidence scores, i.e., we measure $\frac1T\sum_{t=1}^{T}C_{\theta}(r_{t}|x, r_{1},\ldots, r_{t-1})$ for multiple statements in the reasoning path $r=[r_1,\ldots,r_T]$.\footnote{We compare both measures in \Cref{subsec:abb}.} We also incorporate the answer-level confidence score $C_{\theta}(a|x,r)$, measured by $\text{P(True)}$ given the question $x$ and the reasoning path $r$. The implementations and prompts are described in \Cref{appx:prompt}.

Next, we optimize the LLM $M_{\theta}$ to prefer reasoning–answer pairs with high confidence scores, measured as $C_{\theta}(a,r|x)=C_{\theta}(a|x, r)C_{\theta}(r|x)$. To this end, we use online Direct Preference Optimization \citep[DPO]{guo2024direct}, which samples two or more outputs for a given question and optimizes the LLM to assign higher likelihood to the output with higher confidence scores:
\begin{equation}\label{eq:objective}
\mathcal{L}=\log \sigma\left( \beta \log \frac{M_{\theta}(s^{l}|x)}{M_{\text{ref}}(s^{l}|x)}-\beta \log \frac{M_{\theta}(s^{w}|x)}{M_{\text{ref}}(s^{w}|x)}\right)
\end{equation}
where $s^w = [y^{w}, r^{w}]$ denotes a sequence with a higher confidence score than another sequence~$s^l$, $\sigma$ denotes the logistic function, and $\beta$ is a hyperparameter in the DPO. The base reference model $M_{\text{ref}}$ initializes $M_{\theta}$. We fix the reference model during training without updates.

\begin{table*}[t]
\centering
\scalebox{0.95}{
{\begin{tabular}{cccccc}
\toprule
\textbf{Fine-tuning} & \textbf{Decoding} & GSM8K & ARC-Challenge & QPQA$^{\text{ext}}$ & MATH$^{\text{lv5}}$ \\
\midrule
\midrule
\multirow{4}{*}{Not Applied} & Greedy & 90.0 & 89.1 & 30.6 & 45.4 \\
&  \cll Linguistic &  \cll 91.1 & \cll 89.9 & \cll 31.4 & \cll 47.9 \\
& \cll SC & \cll 92.0 & \cll 91.5 & \cll 33.7 & \cll 54.6 \\
& \cll $\text{P(True}|r,a\text{)}$ & \cll 93.2 & \cll 91.3 & \cll 34.1 & \cll 55.0\\
\hline
\rowcolor{gray!20}\multicolumn{6}{l}{\textit{Learning from linguistic self-assessment of answer quality}} \\
\multirow{2}{*}{SR-PO \citep{kumar2024self-rewarding}} & Greedy & 90.9 & 90.2 & 32.6 & 48.1 \\
& \cll Linguistic & \cll 92.6 & \cll 91.3 & \cll 35.8 & \cll 49.3 \\
\rowcolor{gray!20}\multicolumn{6}{l}{\textit{Learning to prefer reasoning with high answer-level confidence $C_{{\theta}}(a|x)$}} \\
\multirow{2}{*}{SC-PO \citep{prasad2024self}} & Greedy & 91.0 & 91.0 & 34.3 & \textbf{49.6} \\
& \cll SC & \cll 93.0 & \cll 92.0 & \cll 36.3 & \cll {55.7} \\
\hline
\rowcolor{gray!20}\rowcolor{gray!20}\multicolumn{6}{l}{\textit{Learning to prefer reasoning with high reasoning-level confidence $C_{{\theta}}(r|x)$ \textbf{(ours)}}} \\
\multirow{2}{*}{\textbf{CORE-PO} \textbf{(ours)}} & Greedy & \textbf{91.3} & \textbf{92.2} & \textbf{37.5} & \textbf{49.6} \\
& \cll $\text{P(True}|r,a\text{)}$ & \cll \textbf{93.5} & \cll \textbf{92.8} & \cll \textbf{38.5} & \cll \textbf{55.8} \\
\bottomrule
\end{tabular}
}}
\caption{\textbf{Performance of self-training methods on Qwen2.5-7B-Instruct.} \textbf{Bold} indicates the best performance under greedy decoding or \colorbox{blue!5}{inference-time scaling methods} (with sampling of eight outputs). Our method outperforms the considered baselines when applying both greedy decoding and inference-time scaling method.} 
\label{tab:qwen}
\end{table*}

\section{Experiments} \label{sec:results}

In this section, we conduct experiments to validate \textbf{CORE-PO} across various reasoning tasks.

\subsection{Experimental setup}


In experiments, we consider two LLMs, Llama3.1-8B-Instruct \citep{meta2024llama3} and Qwen2.5-7B-Instruct \citep{qwen2024qwen25}, as base LLMs to implement our self-training method and baselines.

\noindent \textbf{Tasks and datasets.} We evaluate our self-training method on a range of reasoning tasks using the following datasets:
\begin{itemize}[topsep=1.0pt,itemsep=1.0pt,leftmargin=3.5mm]
\item \textbf{GSM8K} \citep{cobbe2021training} consists of basic math questions requiring multi-step arithmetic reasoning. We use questions in training split for self-training and evaluate on the test split by the accuracy of the generated numerical answers.
\item \textbf{ARC-Challenge} \citep{clark2018think} contains multiple-choice science questions requiring commonsense reasoning. We use questions in the training split for self-training and evaluate on the test split by the accuracy of the selected choices.
\item \textbf{GPQA} \citep{rein2023gpqa} contains graduate-level multiple-choice questions requiring advanced scientific reasoning. We use questions in GPQA-main and GPQA-extended splits for training and evaluation, respectively.
\item \textbf{MATH} \citep{hendrycksmath2021} is a mathematic reasoning benchmark, which consists of challenging high-school math problems. We use questions in the training split for self-training and evaluate on Level-5 questions in the test split.
\end{itemize}
In addition, we evaluate the out-of-domain generalization capabilities of self-trained LLMs using the following two benchmarks:
\begin{itemize}[topsep=-1.0pt,itemsep=1.0pt,leftmargin=3.5mm]
\item \textbf{CRUX} \citep{pmlr-v235-gu24c} is a benchmark for evaluating code understanding and execution. We use tasks of predicting the output of Python functions given inputs, i.e., CRUX$^{\text{out}}$.
\item \textbf{Game of 24} \citep{lile2025game24} is a reasoning benchmark, where the goal is to determine whether a given set of four integers can be combined using operations (addition, subtraction, multiplication, or division) to induce the number $24$.
\vspace{.02in}
\end{itemize}
We provide detailed data statistics of the above datasets in \Cref{appx:datastat}.

\begin{table*}[t]
\centering
\scalebox{0.95}{\begin{tabular}{cccccc}
\toprule
 & & \multicolumn{2}{c}{GSM8K} & \multicolumn{2}{c}{ARC-Challenge}\\
\cmidrule{3-4} \cmidrule{5-6}
\textbf{Fine-tuning} & \textbf{Decoding} & Conf. Score & Reason Acc. & Conf. Score & Reason Acc. \\
\midrule
\midrule
\multirow{2}{*}{Not Applied} & Greedy & 0.89 & 84.2 & 0.84 & 79.2 \\
& \cll $\text{P(True}|\text{r,a)}$ &\cll  0.98 & \cll 89.7 & \cll 0.94 & \cll 81.2 \\
 \midrule
\multirow{2}{*}{\textbf{CORE-PO (ours)}}& Greedy & \textbf{0.94} & \textbf{86.8} & \textbf{0.95} & \textbf{81.5} \\
& \cll $\text{P(True}|\text{r,a)}$ & \cll \textbf{0.99} & \cll \textbf{90.4} & \cll \textbf{0.99} & \cll \textbf{84.9} \\
\bottomrule
\end{tabular}
}
\caption{\textbf{Reasoning-level confidence and accuracy.} The base LLM is Llama3.1-8B-Instruct. \textbf{Bold} indicates the best performance under greedy decoding or \colorbox{blue!5}{inference-time scaling methods} (with sampling of eight outputs). Our self-training method enables the LLM to improve reasoning-level confidence and accuracy.}
\label{tab:reasoning_acc}
\end{table*}

\noindent \textbf{Baselines.} We compare our CORE-PO with existing self-training approaches. We first consider self-rewarding-based preference optimization~\citep[SR-PO]{kumar2024self-rewarding}, which trains LLMs using linguistic self-assessments of answer quality, e.g., assigning higher scores to outputs involving expert-level knowledge. Next, we consider self-consistency preference optimization \citep[SC-PO]{zhang2024self}, which trains LLMs to prefer outputs with high answer-level confidence scores $C_{\theta}(a|x)$ measured via self-consistency score \citep[SC]{wang2023selfconsistency}, i.e., majority voting over multiple sampled outputs.

To compare the performance, we generate outputs from each fine-tuned and base LLM using two decoding schemes: (1) greedy decoding and (2) inference-time scaling methods. For (2), we generate multiple outputs and select the most promising one from the self-assessment score associated with each self-training method: the linguistic self-assessment score (Linguistic) for SR-PO, the majority-voting score over multiple answers (SC) for SC-PO, and the estimated confidence score on the reasoning-answer pair ($\text{P(True}|r,a\text{)}$) for our self-training method.

\noindent \textbf{Implementations.} In our experiments, we apply self-training methods to LLMs using unified training question sets from the aforementioned datasets. Here, we implement our self-training method using \text{monolithic $\text{P(True)}$} to measure confidence scores on reasoning paths.\footnote{We present the results of self-training with statement-wise $\text{P(True)}$ in \Cref{subsec:abb}.} We then evaluate the self-trained LLMs on test question sets from both in-distribution and out-of-distribution datasets. For training, we generate $N=5$ outputs for each question with $T=1.0$ temperature and top-$p=0.9$ \citep{Holtzman2020The}. We also apply a low rank adaptation \citep{hu2022lora} with rank $128$ and $\alpha = 256$ to both Llama3.1–8B–Instruct and Qwen2.5–7B–Instruct. We also use $\beta=0.1$ in direct preference optimization. To apply inference-time scaling methods, we randomly sample eight outputs with $T = 0.7$ and top-$p = 0.9$. We further describe detailed experimental settings and prompts in \Cref{appx:setting}. 

\begin{table}[t]
\centering
\scalebox{0.9}{
{\begin{tabular}{cccc}
\toprule
\textbf{Fine-tuning} & \textbf{Decoding} & CRUX$^{\text{out}}$ & Game of 24  \\
\midrule
\midrule
\multirow{4}{*}{Not Applied} & Greedy & 34.8 &  7.2 \\
& \cll Linguistic & \cll 34.7 & \cll 13.2  \\
& \cll SC & \cll 46.1 & \cll 15.3 \\
& \cll $\text{P(True}|\text{r,a)}$ & \cll 41.0 & \cll 21.0 \\ 
\midrule
\multirow{2}{*}{SR-PO} & Greedy & 31.6 & 8.8  \\ 
& \cll Linguistic & \cll 36.2 & \cll 10.5\\
\midrule
\multirow{2}{*}{SC-PO} & Greedy & 43.8 & 8.3 \\
& \cll SC & \cll \textbf{50.0} & \cll 11.9 \\
\midrule
\multirow{2}{*}{\textbf{CORE-PO}} & Greedy & \textbf{47.1} & \textbf{18.8}\\
& \cll $\text{P(True}|\text{r,a)}$ & \cll {48.0} & \cll \textbf{22.1}\\
\bottomrule
\end{tabular}
}}
\caption{\textbf{Performance of self-training methods on out-of-distribution tasks.} The base LLM is Llama3.1-8B-Instruct. \textbf{Bold} indicates the best performance under greedy decoding or \colorbox{blue!5}{inference-time scaling methods} (with sampling of eight outputs). Our self-training method shows competitive performance or yields best results compared to considered baselines.} 
\label{tab:ood}
\vspace{-.1in}
\end{table}

\subsection{Main results}

\textbf{Competitive performance for self-training.} We present the results in \Cref{tab:llama,tab:qwen}, which are obtained from our implementations using Llama3.1-8B-Instruct and Qwen2.5-7B-Instruct, respectively. One can observe that our self-trained LLMs outperform the base LLMs and the LLMs self-trained with SR-PO and SC-PO algorithms. In particular, one can observe that self-training with reasoning-level confidence (CORE-PO) yields larger gains over self-training with answer-level confidence (SC-PO) on the ARC-Challenge and GPQA benchmarks, whose multiple-choice format often allows incorrect reasoning paths to lead to high-confidence answers by chance. Furthermore, on these benchmarks, one can see that our fine-tuned LLMs achieve higher accuracy with greedy decoding than the base LLMs using inference-time scaling. These overall results highlight that our self-training method enables the LLMs to generate higher-quality answers.

\begin{table*}[t]
\centering
\scalebox{0.95}{\begin{tabular}{ccccc}
\toprule
\textbf{Fine-tuning} & GSM8K & ARC-Challenge & GPQA$^{\text{ext}}$ & MATH$^{\text{Lv5}}$ \\
\midrule
\midrule
Not Applied & 84.2 & 84.5 & 32.4 & 22.6 \\
 \midrule
CORE-PO w/ Monolithic $\text{P(True)}$ &  86.8 & 87.5 & 35.5 & 24.7 \\
CORE-PO w/ Statement-wise $\text{P(True)}$ & 88.5 & 88.0 & 34.1 & 25.3 \\
\bottomrule
\end{tabular}
}
\vspace{-.02in}
\caption{\textbf{Self-training with two different confidence measures.} The base LLM is Llama3.1-8B-Instruct. We use greedy decoding for the comparison. Both monolithic $\text{P(True)}$ and statement-wise $\text{P(True)}$ consistently improve the base model, but neither variant shows a clear advantage over the other.} 
\label{tab:mono_vs_state}
\end{table*}

\begin{table*}[t]

\centering
\scalebox{0.95}{\begin{tabular}{cccc}
\toprule
\textbf{Reward signal} & \textbf{Decoding} & Answer Acc. & Reason Acc. \\
\hline
\hline
\rowcolor{gray!20}\rowcolor{gray!20}\multicolumn{4}{l}{\textit{Learning to prefer reasoning paths leading to correct answers}} \\
\multirow{2}{*}{Answer Acc.}  & Greedy & 87.4 & 73.6 \\
  & \cll $\text{P(True}|\text{r,a)}$ & \cll 88.1 & \cll 78.1 \\
  \hline
\rowcolor{gray!20}\rowcolor{gray!20}\multicolumn{4}{l}{\textit{Learning to prefer high-confidence reasoning paths leading to correct answers \textbf{(ours)}}} \\
\multirow{2}{*}{\textbf{Answer Acc. + Reason Conf. (ours)}} & Greedy & \textbf{88.3} & \textbf{81.9} \\
 & \cll $\text{P(True}|\text{r,a)}$ & \cll \textbf{90.1} & \cll \textbf{85.6} \\
\bottomrule
\end{tabular}
}
\vspace{-.02in}
\caption{\textbf{Experiments with ground-truth answers.} The base LLM is Llama3.1-8B-Instruct. \textbf{Bold} indicates the best performance under greedy decoding or \colorbox{blue!5}{inference-time scaling methods} (with sampling of eight outputs). We use the training and test splits of the ARC-Challenge dataset~\citep{clark2018think}. Incorporating reasoning-level confidence $C_{\theta}(r|x)$ leads to improvements in both answer-level and reasoning-level accuracy.}
\label{tab:ground}
\vspace{-.1in}
\end{table*}

\noindent \textbf{Improved reasoning-level accuracy and confidence in reasoning.} In addition, we also measure the confidence score on the generated reasoning paths and the reasoning-level accuracy.\footnote{We evaluate the reasoning-level accuracy using external oracle \texttt{o4-mini-2025-04-16} \citep{openai2025o4mini} by querying whether the reasoning is correct.} We present the results in \Cref{tab:reasoning_acc}. Here, one can observe that our fine-tuned model increases reasoning-level confidence compared to the base LLM, which coincides with an increase in reasoning-level accuracy. These results also provide evidence that our self-training method enhances the reasoning capabilities of the base LLM by preferring reasoning paths with higher confidence.

\noindent \textbf{Generalization to out-of-distribution tasks.} We also conduct validation on out-of-distribution tasks. We present the results in \Cref{tab:ood}. One can observe that our method yields significant improvements on both CRUX$^{\text{out}}$ and Game of 24, compared to the base LLM. In particular, one can see that our self-training method yields the best performance on Game of 24 when applying both greedy decoding and inference-time scaling. While SC-PO shows competitive performance on CRUX$^{\text{out}}$, it shows limited improvement on Game of 24. These results highlight that our method generalizes better to out-of-distribution tasks than existing baselines.

\subsection{Ablation studies}\label{subsec:abb}

\noindent \textbf{Monolithic $\text{P(True)}$ vs. statement-wise $\text{P(True)}$.} We also conduct experiments by implementing our method using \text{statement-wise $\text{P(True)}$}. We present the results in \Cref{tab:mono_vs_state}. Here, one can see that neither method consistently outperforms the other: self-training with \text{statement-wise $\text{P(True)}$} yields high performance on the GSM8k, ARC-Challenge, and MATH$^{\text{lv5}}$ benchmarks, but yields relatively low performance on the GPQA$^{\text{ext}}$ benchmark. Nevertheless, both methods consistently outperform the base LLM. These results highlight that the performance improvements of our method do not stem from a particular implementation of confidence estimation, but from the philosophy of preferring high reasoning-level confidence.

\noindent \textbf{Fine-tuning with ground-truth answers.} We also conduct experiments by incorporating reasoning-level confidence into a conventional fine-tuning (not self-training) method, which uses ground-truth answers and defines answer-level accuracy as the reward \citep{zelikman2022star,trung-etal-2024-reft,deepseekai2025deepseekr1incentivizingreasoningcapability}. This experiment is motivated by the following pitfall in this method: rewarding incorrect reasoning paths that yield correct answers by chance. Motivated by this issue, we hypothesize that incorporating reasoning-level confidence can prevent the LLM from preferring incorrect reasoning paths that lead to correct answers. Specifically, we compare two fine-tuning methods: one that assigns rewards based solely on answer-level accuracy, and another that incorporates both answer-level accuracy and reasoning-level confidence.

In \Cref{tab:ground}, we present the results of experiments conducted on the multiple-choice questions from the ARC-Challenge dataset. One can see that incorporating reasoning-level confidence improves both answer-level and reasoning-level accuracies. Notably, the LLM fine-tuned solely on answer-level accuracy shows significantly lower reasoning-level accuracy compared to its answer-level accuracy. We observe that this degradation stems from the model's reasoning patterns, which tend to generate unnecessarily long reasoning paths by evaluating all options in the multiple-choice question. Here, the LLM often makes mistakes in evaluating some of the choices, even when the final answer is correct. \Cref{tab:exex} presents examples of these patterns.

\begin{table}[t]
\vspace{0.05in}
\centering
\renewcommand{\arraystretch}{1.2}
\scalebox{0.78}{\begin{tabular}{|p{9cm}|}
\hline
\textbf{Question} \\
\hline
At which temperature does water freeze? \\ 
{[I] 0°C, [II] 32°C, [III] 100°C, [IV] 212°C} \\
\hline
\textbf{Fine-tuning with Answer Acc.} \\
\hline
[I] is correct: water freezes at 0°C. 
[II] is incorrect: \rred{32 is boiling point in Fahrenheit.}
[III] is incorrect: 100°C is the boiling point of water. 
[IV] is incorrect: 212 is the boiling point in Fahrenheit. 
\textbf{Answer: [I]} \\
\hline
\textbf{Fine-tuning with Answer Acc. + Reason Conf. (ours)} \\
\hline
The question asks for the freezing point of water. In the Celsius scale, this is 0°C, a well-known scientific fact across disciplines such as chemistry and physics. \textbf{Answer: [I]} \\
\hline
\end{tabular}}
\vspace{-.05in}
\caption{\textbf{Example of generated reasoning paths.} The reasoning content is summarized due to its excessive length. The first reasoning path involves \rred{errors} despite leading to the correct answer.}
\label{tab:exex}
\vspace{-.15in}
\end{table}

\section{Conclusion} \label{sec:conclusion}

In this paper, we propose a new confidence-based self-training method that addresses a key limitation of existing approaches: the exclusive reliance on answer-level confidence, which does not capture the overall quality of the reasoning. By incorporating reasoning-level confidence, our method fine-tunes LLMs to prefer high-confidence reasoning paths with fewer errors, thereby improving their reasoning capabilities. Empirical results on six benchmarks show that our method improves the reasoning capabilities of LLMs on both in-distribution and out-of-distribution tasks, outperforming existing self-training methods.

\section*{Limitations}


\noindent \textbf{Confidence measures.} Although we use a confidence measure that is relatively reliable than pure likelihoods over generated sequences, it can suffer from overconfidence due to the inherent calibration issues of large language models. This still poses the risk of reinforcing incorrect reasoning paths that are assigned high confidence scores due to miscalibrated confidence estimates. Next, our evaluation considers confidence metrics based solely on $\text{P(True)}$, but incorporating alternative measures, e.g., semantic entropy \citep{kuhnsemantic} or contextualized likelihood \citep{lin-etal-2024-contextualized}, may provide a more robust estimation of the confidence. An interesting avenue for future work is to develop and incorporate more robust and well-calibrated confidence measures into our method.

\noindent \textbf{Language-specific experiments.} Our experiment focuses exclusively on English, and we do not explore the applicability of our method to other languages, e.g., morphologically rich or typologically diverse languages. Since reasoning pattern and confidence calibration can vary significantly across languages due to linguistic structure and pretraining data distribution, it remains unexplored whether our findings generalize beyond English.

\noindent \textbf{Prompting.} We evaluate our method in a zero-shot setting using the default system prompt, i.e., ``Be a helpful assistant.''. However, more advanced prompting strategies, such as few-shot prompting or task-specific system prompts \citep{brown2020language}, may further improve performance.

\noindent \textbf{Human evaluation.} In this paper, we do not conduct human evaluation to assess the quality or faithfulness of the generated outputs, leaving open the question of alignment with human judgment. 

\noindent \textbf{Experiments on larger-scale models.} Our experiments only consider large language models with up to 7.5B or 8B parameters due to the limited computational budgets. The generalizability of our method to larger models (e.g., 70B) remains unexplored and is left for future work.

\section*{Acknowledgments}

This work was partly supported by Institute for Information \& communications Technology Planning \& Evaluation(IITP) grant funded by the Korea government(MSIT) (RS-2019-II190075, Artificial Intelligence Graduate School Support Program(KAIST)), National Research Foundation of Korea(NRF) grant funded by the Ministry of Science and ICT(MSIT) (No. RS-2022-NR072184), GRDC(Global Research Development Center) Cooperative Hub Program through the National Research Foundation of Korea(NRF) grant funded by the Ministry of Science and ICT(MSIT) (No. RS-2024-00436165), the Institute of Information \& Communications Technology Planning \& Evaluation(IITP) grant funded by the Korea government(MSIT) (RS-2025-02304967, AI Star Fellowship(KAIST)), and the Graduate School of Artificial Intelligence (GSAI) Cluster at POSTECH.

\bibliography{ref.bib}

\newpage
\appendix

\section{Related works} \label{sec:related}

\noindent \textbf{Confidence measures for LLMs.} Large language models (LLMs) often generate incorrect outputs due to hallucinations, which highlights the importance of estimating confidence in their outputs. To this end, several methods have been proposed, including self-consistency \citep{wang2023selfconsistency}, semantic entropy over semantically equivalent sequences \citep{kuhnsemantic}, the probability of truth $\text{P(True)}$ \citep{kadavath2022languagemodelsmostlyknow}, or asking the model to express its confidence in linguistic form \citep{tian-etal-2023-just}. In addition, \citet{lin-etal-2024-contextualized} propose computing confidence using the likelihoods of important tokens that determine the semantics of the sequence. Although such confidence measures have mainly been applied to single-statement factuality checks, recent studies have shown that measures based on $\text{P(True)}$ can be utilized to estimate the confidence in the reasoning path \citep{becker2024cyclesthoughtmeasuringllm,taubenfeld2025confidence}.

\noindent \textbf{Inference-time scaling methods for LLMs.} Inference-time scaling methods improve the quality of LLM outputs by self-assessing multiple generated outputs. Among them, the self-consistency-based method \citep{wang2023selfconsistency} selects the most frequent answer obtained through majority voting. Other approaches include tree-of-thoughts \citep{yao2023tree}, which expands the search space over intermediate steps, and self-refinement inference \citep{madaan2023self}, which iteratively refines outputs using the LLM itself. In addition, \citet{taubenfeld2025confidence,wan-etal-2025-reasoning} recently proposed incorporating confidence scores on reasoning paths into the self-consistency method and showed notable performance improvements.

\noindent \textbf{Training of reasoning for LLMs.} To enhance reasoning abilities, LLMs are initially fine-tuned using various supervision signals. A straightforward approach is supervised fine-tuning on high-quality reasoning datasets \citep{cobbe2021training,trung-etal-2024-reft}, or direct preference optimization on reasoning datasets annotated with human preferences \citep{meta2024llama3}. While effective, collecting such datasets is costly. As an alternative approach, several studies instead consider reinforcement learning methods that rely solely on ground-truth answers, using answer-level accuracy as the reward signal \citep{zelikman2022star,trung-etal-2024-reft,deepseekai2025deepseekr1incentivizingreasoningcapability}. In addition, several studies propose training process reward models (PRMs) that assess the quality of intermediate reasoning steps. \citet{lightman2024lets} and \citet{jiao2024learning} train PRMs using human-annotated preferences on individual reasoning steps and answer-level ground-truth labels, respectively.

\noindent \textbf{Other self-training approaches for LLMs.} We further describe self-training methods derived from inference-time scaling techniques that are not based on confidence. First, \citet{kumar2024self-rewarding} propose using linguistic assessments of output quality, e.g., assigning high scores to outputs that exhibit expert-level knowledge, as reward signals for fine-tuning LLMs. Next, \citet{ranaldi2024self} propose using outputs obtained from self-refinement inference \citep{madaan2023self}, as these outputs typically exhibit higher quality than the initial outputs. Lastly, \citet{zhang2024chain} leverage preference signals over intermediate reasoning steps derived from tree-of-thoughts inference \citep{yao2023tree}. A concurrent line of work, Absolute Zero \citep{zhao2025absolutezeroreinforcedselfplay} shows that LLMs can improve their reasoning abilities by self-generating and solving code-based tasks, without relying on external data or human supervision.

\section{Experimental details}\label{appx:setting}

We use all datasets and models in accordance with their intended use for academic research, following their respective licenses.

\subsection{Observational experiment}\label{appx:observe}

For the observational experiment, we use questions from the GPQA-main dataset \citep{rein2023gpqa}. For each question, we generate an output consisting of a reasoning path and a final answer using Llama3.1-8B-Instruct \citep{meta2024llama3}. To be specific, to obtain reasoning paths with high answer-level or reasoning-level confidence, we generate $16$ outputs per question and select the one with the highest confidence score. As a result, we obtain a triplet (question, reasoning path with high answer-level confidence, reasoning path with high reasoning-level confidence) for each question in the GPQA-main dataset. We then evaluate the correctness of each reasoning path using an external tool: \texttt{o4-mini-2025-04-16}. Before evaluating the correctness of each reasoning path using \texttt{o4-mini-2025-04-16}, we first consider the reasoning to be incorrect if the answer is incorrect.

\newpage 

\subsection{Prompts}\label{appx:prompt}

\textbf{Prompt for solving ARC-Challenge and GPQA.} We use the following prompt to solve the given multiple-choice question \textcolor{blue}{[question]}.

\begin{table}[H]
\centering
\scalebox{0.77}{
\begin{tcolorbox}[
    colback=gray!5!white,
    colframe=gray!60!black,
    width=1.3\linewidth,
    boxsep=5pt,
    left=5pt, right=5pt,
    before upper=\raggedright,  
]
Answer the following question using **reasoning** before providing a final answer. Provide a precise, structured, and well-reasoned response.\\
\medskip
\medskip
**Question:** \textcolor{blue}{[question]} \\
\medskip
\medskip
\#\#\# Response Format \\
\medskip
**Understanding the question:** <identify key details> \\
\medskip
**Reasoning:** <perform chain-of-thought> \\
\medskip
**Final answer:** ``The answer is <choose the most promising single answer from [I] / [II] / [III] / [IV]> which is <copy the content>'' \\
\medskip
\medskip
Ensure correctness and clarity. Return a concise and definitive response to the question. DO NOT RETURN TWO OR MORE ANSWERS. STRICTLY FOLLOW THE RESPONSE FORMAT.
\end{tcolorbox}}
\end{table}

\noindent \textbf{Prompt for solving GSM8K and MATH.} We use the following prompt to solve the given numeric-response question \textcolor{blue}{[question]}.

\begin{table}[H]
\centering
\scalebox{0.77}{
\begin{tcolorbox}[
    colback=gray!5!white,
    colframe=gray!60!black,
    width=1.3\linewidth,
    boxsep=5pt,
    left=5pt, right=5pt,
    before upper=\raggedright,  
]
Answer the following question using **reasoning** before providing a final answer. Provide a precise, structured, and well-reasoned response.\\
\medskip
\medskip
**Question:** \textcolor{blue}{[question]} \\
\medskip
\medskip
\#\#\# Response Format \\
\medskip
**Understanding the question:** <identify key details> \\
\medskip
**Reasoning:** <perform chain-of-thought> \\
\medskip
**Final answer:** ``The answer is \$<value>\$'' \\
\medskip
\medskip
Ensure correctness and clarity. Return a concise and definitive response to the question. STRICTLY FOLLOW THE RESPONSE FORMAT.
\end{tcolorbox}}
\end{table}

\newpage

\noindent \textbf{Prompts for solving questions in Game of 24.} We use the following prompt to complete the expression given the four digit numbers \textcolor{blue}{[four digits]}.

\begin{table}[H]
\centering
\scalebox{0.77}{
\begin{tcolorbox}[
    colback=gray!5!white,
    colframe=gray!60!black,
    width=1.3\linewidth,
    boxsep=5pt,
    left=5pt, right=5pt,
    before upper=\raggedright,  
]
Answer the following question using **reasoning** before providing a final answer. Provide a precise, structured, and well-reasoned response.\\
\medskip
\medskip
**Question:** "Write an equation using basic arithmetic operations (+ - * /) to obtain \$24\$ from the four given numbers, e.g., ``(4 + 8) * (6 - 4) = 24'' from the input [4, 4, 6, 8]. You must use all the given numbers exactly once, i.e., simply rearrange them. Do not use any additional numbers. Parentheses can be used to control the order of operations. 
Now, write an expression using exactly the given numbers \textcolor{blue}{[four digits]} that results in \$24\$.\\
\medskip
\medskip
\#\#\# Response Format \\
\medskip
**Understanding the question:** <identify key details> \\
\medskip
**Reasoning:** <perform chain-of-thought> \\
\medskip
**Final answer:** ``The answer is \$<value>\$'' \\
\medskip
\medskip
Ensure correctness and clarity. Return a concise and definitive response to the question. THE LHS EXPRESSION MUST USE THE FOUR GIVEN NUMBERS EXACTLY ONCE. DO NOT SIMPLIFY THE FINAL EQUATION. STRICTLY FOLLOW THE RESPONSE FORMAT.
\end{tcolorbox}}
\end{table}

\noindent \textbf{Prompt for solving Crux$^{\text{out}}$.} We use the following prompt to predict the output given the code \textcolor{blue}{[code]} and the input \textcolor{blue}{[input]}.

\begin{table}[H]
\centering
\scalebox{0.77}{
\begin{tcolorbox}[
    colback=gray!5!white,
    colframe=gray!60!black,
    width=1.3\linewidth,
    boxsep=5pt,
    left=5pt, right=5pt,
    before upper=\raggedright,  
]
You are given a Python function and an assertion containing an input to the function. Complete the assertion with a literal (no unsimplified expressions, no function calls) containing the output when executing the provided code on the given input, even if the function is incorrect or incomplete. Execute the program step by step as **reasoning** before providing a final answer. Provide a precise, structured, and well-reasoned response. \\
\medskip
**Code:** \textcolor{blue}{[code]} \\
\medskip
\medskip
\#\#\# Response Format \\
\medskip
**Reasoning:** <perform chain-of-thought (step-by-step execution)> \\
\medskip
**Final answer:** assert f(\textcolor{blue}{[input]}) == <output>""" \\
\medskip
\medskip
Ensure correctness and clarity. Return a concise and definitive response to the question. STRICTLY FOLLOW THE RESPONSE FORMAT.
\end{tcolorbox}}
\end{table}

\newpage

\noindent \textbf{Prompt for confidence estimation.} We use the following prompts to estimate the confidence score on reasoning path \textcolor{blue}{[reasoning]} and answer \textcolor{blue}{[answer]}. As we obtain multiple reasoning paths in inference-time scaling, we also augment additional reasoning paths \textcolor{blue}{[example $i$]} for $i = 1, \ldots, M$ in estimating the confidence score on the reasoning path \textcolor{blue}{[reasoning]}, where $M = 4$.

\begin{table}[H]
\centering
\scalebox{0.77}{
\begin{tcolorbox}[
    colback=gray!5!white,
    colframe=gray!60!black,
    width=1.3\linewidth,
    boxsep=5pt,
    left=5pt, right=5pt,
    title=Measuring monolithic $C_{\theta}(r|x)$,
    before upper=\raggedright, 
]
Answer whether the **selected reasoning** is correct for the given **question**. Additionally, we provide randomly generated reasoning before presenting the selected reasoning. \\
\medskip
\medskip
**Question:** \textcolor{blue}{[question]} \\
\medskip
\medskip
**Randomly generated reasoning 1 (this may be either correct or incorrect):** \textcolor{blue}{[example $1$]} \\
\medskip
$\cdots$ \\
\medskip
**Randomly generated reasoning M (this may be either correct or incorrect):** \textcolor{blue}{[example $M$]} \\
\medskip
\medskip
**Selected reasoning:** \textcolor{blue}{[reasoning]}\\
\medskip
\medskip
Is the **selected reasoning** correct? \\
\medskip 
A) True \\ 
\medskip 
B) False \\ 
\medskip
The **selected reasoning** is: [A / B, depending on whether the **selected reasoning** is correct given the **question**]
\end{tcolorbox}}
\end{table}

\begin{table}[H]
\centering
\scalebox{0.77}{
\begin{tcolorbox}[
    colback=gray!5!white,
    colframe=gray!60!black,
    width=1.3\linewidth,
    boxsep=5pt,
    left=5pt, right=5pt,
    title=Measuring statement-wise $C_{\theta}(r_k|x{,r_1,\ldots,r_{k-1}})$,
    before upper=\raggedright, 
]
Answer whether the **new reasoning statement** is correct for the given **previous reasoning statements** and the **question**.\\
\medskip
\medskip
**Question:** \textcolor{blue}{[question]} \\
\medskip
\medskip
**Previous reasoning statements:** \\
\textcolor{blue}{[step-$1$]} \\
$\cdots$ \\
\textcolor{blue}{[step-$(k-1)$]} \\
\medskip
\medskip
**New reasoning statement:** \textcolor{blue}{[step-$k$]} \\
\medskip
\medskip
Is the **new reasoning statement** correct? \\
\medskip 
A) True \\ 
\medskip 
B) False \\ 
\medskip
The **new reasoning statement** is: [A / B, depending on whether the **new reasoning statement** is correct given the **previous reasoning statements** and the **question**]
\end{tcolorbox}}
\end{table}

\begin{table}[H]
\centering
\scalebox{0.77}{
\begin{tcolorbox}[
    colback=gray!5!white,
    colframe=gray!60!black,
    width=1.3\linewidth,
    boxsep=5pt,
    left=5pt, right=5pt,
    title=Measuring $C_{\theta}(a|{r,x})$,
    before upper=\raggedright, 
]
Answer whether the **selected answer** is correct for the given **question**, based on the provided **reasoning**.\\
\medskip
\medskip
**Question:** \textcolor{blue}{[question]} \\
\medskip
\medskip
**Reasoning:** \textcolor{blue}{[reasoning]} \\
\medskip
\medskip
**Selected answer:** \textcolor{blue}{[answer]} \\
\medskip
\medskip
Is the **selected answer** correct? \\
\medskip 
A) True \\ 
\medskip 
B) False \\ 
\medskip
The **selected answer** is: [A / B, depending on whether the **selected answer** is correct given the **question** and the **reasoning**]
\end{tcolorbox}}
\end{table}

\noindent \textbf{Evaluating reasoning correctness.} We use the following prompt to evaluate the reasoning \textcolor{blue}{[reasoning]} given the question \textcolor{blue}{[question]}.

\begin{table}[H]
\centering
\scalebox{0.77}{
\begin{tcolorbox}[
    colback=gray!5!white,
    colframe=gray!60!black,
    width=1.3\linewidth,
    boxsep=5pt,
    left=5pt, right=5pt,
    before upper=\raggedright,  
]
Given a question, answer whether the reasoning could be correct. \\
\medskip
Respond ONLY in JSON format:\\
\medskip
\{ \\
"verdict": "correct" or "incorrect" \\
\} \\
\medskip
\medskip
**Question:** \textcolor{blue}{[question]} \\
\medskip
\medskip
**Reasoning:** \textcolor{blue}{[reasoning]} 
\end{tcolorbox}}
\end{table}

\subsection{Data statistics}\label{appx:datastat}

We provide detailed data statistics for the datasets used in training and evaluation. The training and test splits of GSM8k dataset \citep{cobbe2021training} contain $7.4$k and $1.3$k questions, respectively. The ARC-Challenge dataset \citep{clark2018think} includes training, validation, and test splits, containing $1.1$k, $0.3$k, and $1.1$k questions, respectively. For the GPQA dataset \citep{rein2023gpqa} (involving $0.4$k and $0.5$k questions in main and extended splits), we use questions with lengths below $1,280$, where the resulting main and extended splits include $420$ and $509$ questions, respectively. The MATH dataset~\citep{hendrycksmath2021} contains $7.5$k training questions and $0.7$k Level-5 test questions. The CRUXEval \citep{guo2024direct} and Game of 24 dataset \citep{lile2025game24} contain $0.8$k and $1.3$k questions, respectively.\footnote{We further clarify that GSM8K, MATH, 24-Game, and CruxEval are released under open-source licenses (Apache 2.0 or MIT). GPQA and ARC-Challenge are distributed under the CC-BY-4.0 license.}

\subsection{Implementations}\label{appx:imple}

We use four NVIDIA A100 SXM4 80GB GPUs. We save checkpoints every 200 steps and select the model with the highest accuracy on the ARC-Challenge validation split. Training the selected model typically takes two to four days. We apply hyper-parameter searching for learning rate over $\{1\textrm{e-}6,5\textrm{e-}6\}$. We also apply a low rank adaptation \citep{hu2022lora} with rank $128$ and $\alpha = 256$. At each gradient step, gradient clipping with a maximum norm of $1.0$ is applied. We report results from a single run. 
\begin{itemize}[topsep=-1.0pt,itemsep=1.0pt,leftmargin=3.5mm]
\item For CORE-PO (ours), which uses online DPO, we generate $N = 5$ outputs for each question in the training set. We construct preference pairs of outputs by evaluating their confidence measures, as described in \Cref{subsec:our_method}. The detailed prompts are provided in \Cref{appx:prompt}. 
\item For SR-PO \citep{kumar2024self-rewarding} which uses offline DPO, we generate $N=5$ outputs for each question in the training set. Then, we construct preference pairs by evaluating their scores using original self-rewarding prompts of SR-PO. In addition, we consider multiple iterations proposed in this method \citep{kumar2024self-rewarding}, where each iteration involves an update of the reference model in DPO. We conduct two iterations in our experimental setup, where the LLM achieving the highest validation accuracy is selected.
\item For SC-PO \citep{prasad2024self} which uses offline DPO. This method samples a larger number of outputs ($N=8$) for each question, since preference pairs are constructed only when the majority voting scores over answers differ by at least $3$ \citep{prasad2024self}. A smaller number of outputs ($N=5$) often fails to construct preference pairs under this criterion. In addition, we consider multiple iterations proposed in this method \citep{prasad2024self}, where each iteration involves an update of the reference model in DPO. We conduct two iterations in our experimental setup, where the LLM achieving the highest validation accuracy is selected.
\vspace{.05in}
\end{itemize}
We further clarify that our implementations are based on the \texttt{transformers} library \citep{wolf-etal-2020-transformers}, the \texttt{trl} library \citep{vonwerra2022trl}, and the \texttt{accelerate} library \citep{accelerate}.\footnote{We use Qwen2.5-7B (Apache 2.0) and LLaMA 3.1-8B (LLaMA 3.1 Community License), both of which allow use and redistribution under their respective terms.}

\section{Use of AI assistants}

We used AI-based writing assistants to improve the grammar. These tools were
used only for editorial improvements. The technical content, methodology, and experimental results were entirely authored by the researchers.

\end{document}